\documentclass[journal]{IEEEtran}

\ifCLASSINFOpdf
\else
\fi
\usepackage[utf8]{inputenc} 
\usepackage[T1]{fontenc}    
\usepackage{url}            
\usepackage{booktabs}       
\usepackage{amsfonts}       
\usepackage{nicefrac}       
\usepackage{microtype}      
\usepackage{bm}
\usepackage{multirow}
\usepackage{array}
\usepackage{threeparttable}
\usepackage{graphicx}
\usepackage{wrapfig}
\usepackage{amsmath}
\usepackage{color}

\usepackage[colorlinks,
            linkcolor=red,
            anchorcolor=blue,
            citecolor=green
            ]{hyperref}

\begin{document}
%
\title{GTAE: Graph-Transformer based Auto-Encoders for Linguistic-Constrained Text Style Transfer}
%
%
%

\author{Yukai Shi~\textsuperscript{*} , Sen Zhang~\textsuperscript{*} ,Chenxing Zhou, Xiaodan Liang, Xiaojun Yang, Liang Lin
~\thanks{
* The first two authors share equal-authorship.
}

}

\markboth{}%
{Shell \MakeLowercase{\textit{et al.}}: Bare Demo of IEEEtran.cls for IEEE Journals}
%



\maketitle

\begin{abstract}

Non-parallel text style transfer has attracted increasing research interests in recent years. Despite successes in transferring the style based on the encoder-decoder framework, current approaches still lack the ability to preserve the content and even logic of original sentences, mainly due to the large unconstrained model space or too simplified assumptions on latent embedding space. Since language itself is an intelligent product of humans with certain grammars and has a limited rule-based model space by its nature, relieving this problem requires reconciling the model capacity of deep neural networks with the intrinsic model constraints from human linguistic rules. To this end, we propose a method called Graph Transformer based Auto Encoder (GTAE), which models a sentence as a linguistic graph and performs feature extraction and style transfer at the graph level, to maximally retain the content and the linguistic structure of original sentences. Quantitative experiment results on three non-parallel text style transfer tasks show that our model outperforms state-of-the-art methods in content preservation, while achieving comparable performance on transfer accuracy and sentence naturalness.
\end{abstract}


%
\IEEEpeerreviewmaketitle

\section{Introduction}

As one of the important content in social multimedia, language draws considerable attention and produce a series of text-related systems~\cite{zhang2019resumevis,yin2012latent,wang2018concept,thukral2019diffque,cui2019short}. However, the text style has long been neglected in above applications and systems. Althought style transfer technologies have made significant progress in the field of computer vision~\cite{gatys2016image, liu2016coupled, johnson2016perceptual,zhu2017unpaired}, their success in natural language processing is much limited. Recently some preliminary works have emerged aiming at this task for text corpus~\cite{hu2017toward, shen2017style, fu2018style}, where a sentence is transferred to a target style attribute (e.g. sentiment, gender, opinion, etc.) while preserving the same content except for the style part. The manipulation of sentence attributes has wide applications in dialog systems and many other natural language fields. Different from related tasks like machine translation~\cite{bahdanau2015neural} and generic text generation~\cite{wen2017a}, corpus for text style transfer are usually non-parallel, and thus the training process is performed in an unsupervised manner. Lacking paired text corpus poses a great challenge to preserving the style-independent content while transferring the style guided by other datasets. 

Current approaches mainly fall into the encoder-decoder paradigm. An encoder is used to extract latent embeddings of the sentence. Content and style embeddings are then disentangled in this space, so that style transfer can be achieved by manipulating the style feature vector, followed by a decoder to generate style-transferred sentences while keeping the content embeddings unchanged~\cite{hu2017toward}. However, in practice these two components are often deeply entangled and it is difficult to separate them in the latent space empirically. Besides, early variational autoencoder~\cite{kingma2014auto} based methods take Gaussian distribution as the prior of latent features, which is a too simplified assumption for natural language. Attempts to alleviate these problems include generating the prior with a learned neural network to replace the naive Gaussian assumption~\cite{zhao2018adversarially}, adopting a cross-alignment strategy~\cite{shen2017style} and translating the sentence to the embedding space of another language under the assumption that translation will remove style-related features automatically~\cite{logeswaran2018content, prabhumoye2018style, lample2019multiple}. Nevertheless, due to the large searching space of models for natural language processing tasks and the lack of paired training corpus, it is non-trivial to retain semantic information in the learned content features. In practice, these approaches could still suffer from the large model space of natural language and fail to preserve the content and even logic of original sentences.

\begin{figure}
  \centering
  \includegraphics[width=0.50\textwidth]{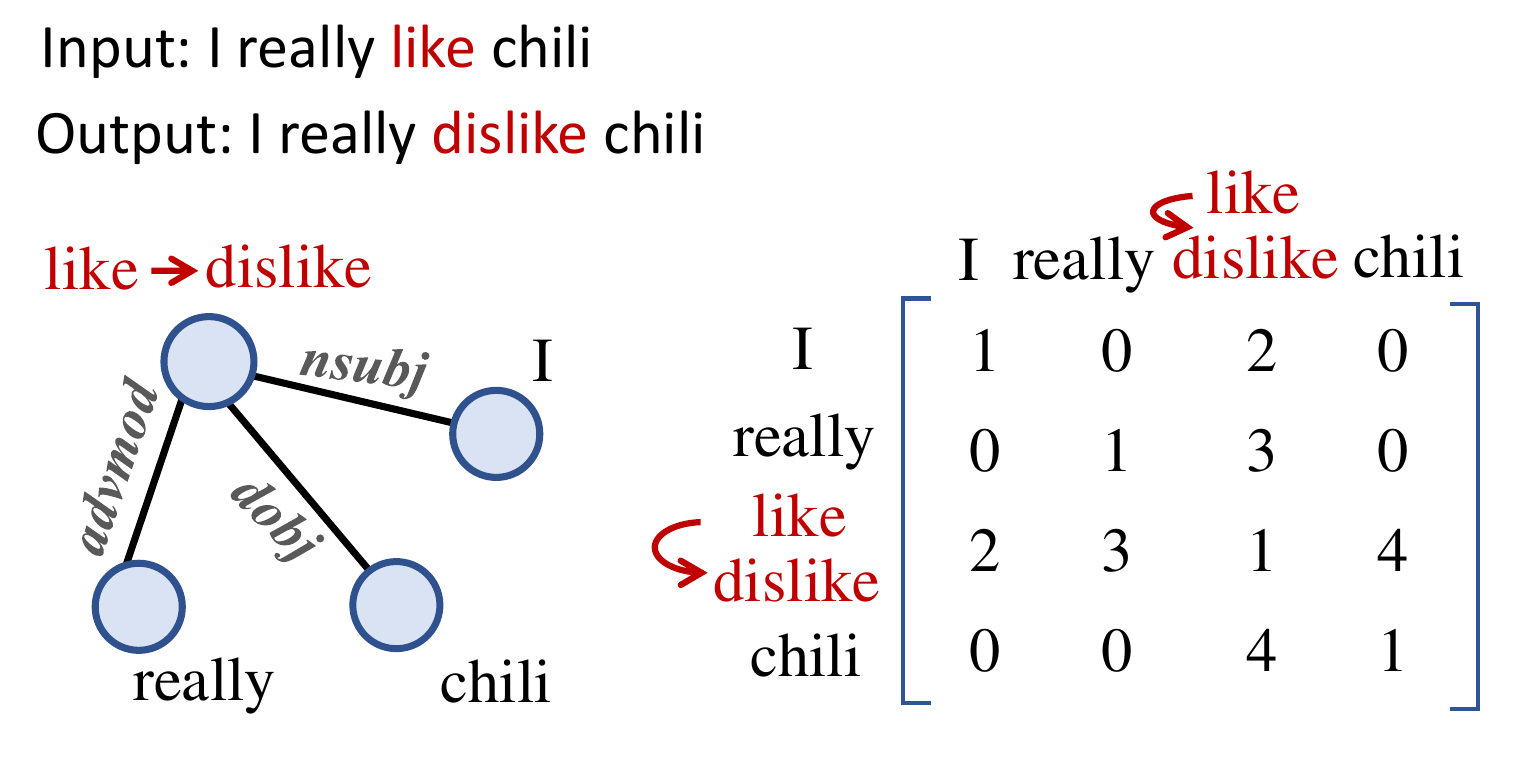}
  \caption{An illustration of the constructed linguistic graph from the dependency tree parser and the style transfer process. The input sentiment is transferred from positive to negative while preserving style-irrelevant contents in the original sentence. Values in the adjacency matrix represent different dependency types determined by grammar rules, which are then used as the imposed linguistic constraints.}
  \label{fig:illustration}
\end{figure}

However, language itself is an intelligent product of human beings, and by its nature should be limited to a much smaller model space based on linguistic rules which make up a language. Without parallel text corpus to restrict the content and the logic of transferred sentences, leveraging the linguistic knowledge provided by language grammars and logic rules could be an effective approach to prevent semantic information loss during the unsupervised training process. Inspired by this observation, we propose Graph-Transformer based Auto Encoder (GTAE) for non-parallel text style transfer, which models the linguistic constraints as graphs explicitly, and performs feature extraction and style transfer at the graph level, so as to maximally retain the content and the linguistic structure of original sentences. {\color{black}{An illustration is provided in Fig.~\ref{fig:illustration}.}} In particular, we design two modules, i.e. a Self Graph Transformer (SGT) that updates the latent node embeddings with the desired style attribute and a Cross Graph Transformer (CGT) that retrieves the semantic information of input sentences, to replace the encoder and the decoder respectively. A linguistic graph is constructed using a public available semantic dependency parser based on learned grammar rules~\cite{manning2014the}. Both SGT and CGT modules perform latent graph transformations constrained by the same linguistic graph structure for each input sentence. Then a simple RNN rephraser is used to generate final sentences from the transferred graphs. 

To evaluate the performance of our proposed method, we conduct experiments on three text style transfer datasets for sentiments, political slants and title types, where our model outperforms state-of-the-art models in terms of content preservation as expected, while achieving highly comparable transfer intensities and naturalness. The main contributions of this work are:

 (1) We propose to restrict the model space of natural language by its linguistic rules, rather than training without constraints or making too simplified assumptions, to mimic the natural language generation process of human. 
 
 (2) We introduce a graph structure to model the linguistic constraints and propose a framework named Graph-Transformer based Auto Encoder to manipulate the latent space at the graph level. And our model is highly extendable for other tasks and model designs.
 
 (3) Empirical studies demonstrate the ability of our text style transfer system in preserving the semantic content. And we show that feature transformation in the graph-constrained model space will not affect transfer intensities and naturalness as well.
\section{Related Work}
\paragraph{Image Style Transfer}
Style transfer has made great progress in computer vision. \cite{gatys2016image} extracts content and style features and then constrains the synthesized image to be close to both the content of source image and the target style. \cite{johnson2016perceptual} further exploits a perceptual loss that achieves better optimization efficiency. Besides, several works have built their models upon the generative adversarial networks~\cite{yi2017dualgan,zhu2017unpaired} for cross-domain image transformation. These works provide insightful analysis and solutions for style transfer, however, it is non-trivial to adapt these methods into the text domain due to the essential difference between image and natural language.

\paragraph{Non-Parallel Text Style Transfer}
Style transfer with non-parallel text corpus has attracted increasing research interests in recent years. \cite{hu2017toward} used variational autoencoder (VAE) to generate sentences with controllable attributes by disentangling latent representations. \cite{fu2018style} further introduced adversarial training to learn style-irrelevant content embeddings. To relieve the simplistic Gaussian prior of latent embeddings in VAE~\cite{kingma2014auto}, \cite{shen2017style} proposed a cross-aligned strategy to direcly match transferred text distributions with two adversarial discriminators, assuming a shared latent content distribution for both styles. {\color{black}{As a pioneer work in text style transfer,~\cite{xu2018unpaired} employs a reinforced model on non-parallel  text data to present style transfer.~\cite{zhang2018style} uses pseudo-parallel data and neural machine translation to achieve language style transfer. \cite{luo2019dual} proposed a dual agent reinforcement algorithm for text style transfer. Similarly, \cite{luo2019towards} incorporates a Gaussian kernel layer to finely control the sentiment intensity to present fine-grained sentiment transfer. IMaT\cite{jin2019imat} constructs a pseudo-parallel data by aligning semantically similar sentences from pair-wise corpora to implement text attribute transfer. \cite{dai2019style} presents a Style Transformer that employs an attention mechanism in Transformer to achieve content preservation and text style transfer.}} 
\cite{zhao2018adversarially} used a generator model to learn the latent priors instead, and minimized the Wasserstain distance between the latent embedding and the learned prior with an adversarial regularization loss. \cite{yang2018unsupervised} replaced binary classifier discriminators with a target domain language model to stabilize training signals from transferred sentences. 

Our work differs from these antecedents by explicitly imposing linguistic constraints into the latent space for better content and logic preservation. It is worth noting that our proposed GTAE is highly extendable with above modeling strategies by simply replacing their encoder-decoder with our SGT-CGT modules. Another line of work resorts to back-translation methods that removes style attributes in the representation space of another language~\cite{logeswaran2018content,prabhumoye2018style,lample2019multiple}. However, their methods still lack effective constraints on the latent space. \cite{li2018delete} argues that text styles are often embodied in certain phrases and proposes a direct delete-and-retrieve approach. Though their method strictly preserve the linguistic structure, the computation load of retrival prohibits its practical use and the hard-manipulation strategy could suffer when style information cannot be simply disentangled from certain phrases. In comparison, GTAE provides a more flexible and computationally efficient solution that allows operations in the latent space and evolve the corresponding style-related parts. {\color{black}{Recently~\cite{tian2018structured} also investigates the incorporation of linguistic constraints into the modeling process. Their work lies on noun consistency for content preservation while ours resorts to dependency parse trees instead, which are complementary to each other.}}

\paragraph{Graph Neural Network}



Graph provides a flexible data representation for many tasks, especially for natural language which contains highly structured syntactic information in spite of its sequential format. Though the concept of graph neural network was proposed much earlier in \cite{gori2005a}, however, graph-level transformations with cutting-edge deep learning techniques are still under-exploration until recent years. Early works attempt to incorporate the success of convolution network into this field, by defining convolutions on groups of node neighbors~\cite{duvenaud2015convolutional,rossi2018interactive,li2019co-saliency,tang2011image}. More recently, research on graph modeling has put more focus on the attention mechanisms~\cite{vaswani2017attention} to enhance the model capacity. \cite{velickovic2018graph} proposed graph attention network to evolve node embeddings by attending to their spatial neighbors. The transformer-based models proposed in \cite{li2019knowledge} and \cite{koncel-kedziorski2019text} are most similar to ours, while our proposed GTAE allows information flow both within the graph itself and cross different graphs. Besides, we further design a learnable global style node in our self-graph transformer to fit the goal of non-parallel text style transfer. 



\section{Graph-Transformer Based Auto-Encoder}

\subsection{Formulation}
Non-parallel text style transfer targets at generating a new sentence $\bm{y}$ with a specified style attribute $\bm{s_y}$ from the conditional distribution $p(\bm y|\bm{x}, \bm{s_y})$ while preserving the same content as the original sentence $\bm{x}$. The difficulty of this task lies in the lack of parallel training corpus due to the high labor cost to transfer sentences manually. Suppose we observe two datasets $\bm{X}=\{\bm{x}^{(i)}|i\in[1,N_x]\}$ and $\bm{Y}=\{\bm{y}^{(j)}|j\in[1,N_y]\}$ with different style attributes $\bm{s_x}$ and $\bm{s_y}$ respectively. Without paired information to provide training loss,  we want to estimate the two distributions $p(\bm{y}|\bm{x}, \bm{s_y})$ and $p(\bm{x}|\bm{y}, \bm{s_x})$. Previous attempts include using auto-encoder based models to reconstruct $\bm{y}$ by manipulating style-embeddings in the latent space. However, these models either make too simplified assumptions on the latent embeddings or fail to constrain the language space effectively, which could be a reason for their unsatisfactory performance of content preservation.

Considering that language by its nature is restricted to a limited feature space due to grammar rules and human logics, we propose a novel framework named Graph Transformer based Auto Encoder (GTAE) to effectively leverage this intrinsic linguistic constraint. We start by giving an overview of this framework. Then we go into details of each component in the next section. First of all, we observe that linguistic structures either extracted by dependency parsers or predicted by information extraction methods can both be sufficiently represented as a graph $G_x=(V,E)$, where the node set $V$ of the linguistic graph contains entities extracted from this sentence, and $E$ is the edge set indicating the relationship between each node pair. Adding linguistic constraints on the latent space thus can be formulated as transforming sentences into the linguistic graphs first, and then perform feature extraction, style transfer, and reconstruction at the graph level:
\begin{eqnarray}
 & p( \bm{y}|\bm{x},   \bm{s_y})=  \int_{G_x}  p(\bm{y}|G_x, \bm{s_y})  p(G_x|\bm{x}) dG_x& \\
  & =  \int_{G'_y}\int_{G_x}  p(\bm{y}|G'_y,G_x)p(G'_y|G_x,\bm{s_y}) p(G_x|\bm{x})dG_xdG'_y.& \label{eq:framework}
\end{eqnarray}

\begin{figure}[t]
  \centering
  \includegraphics[width=0.56\textwidth]{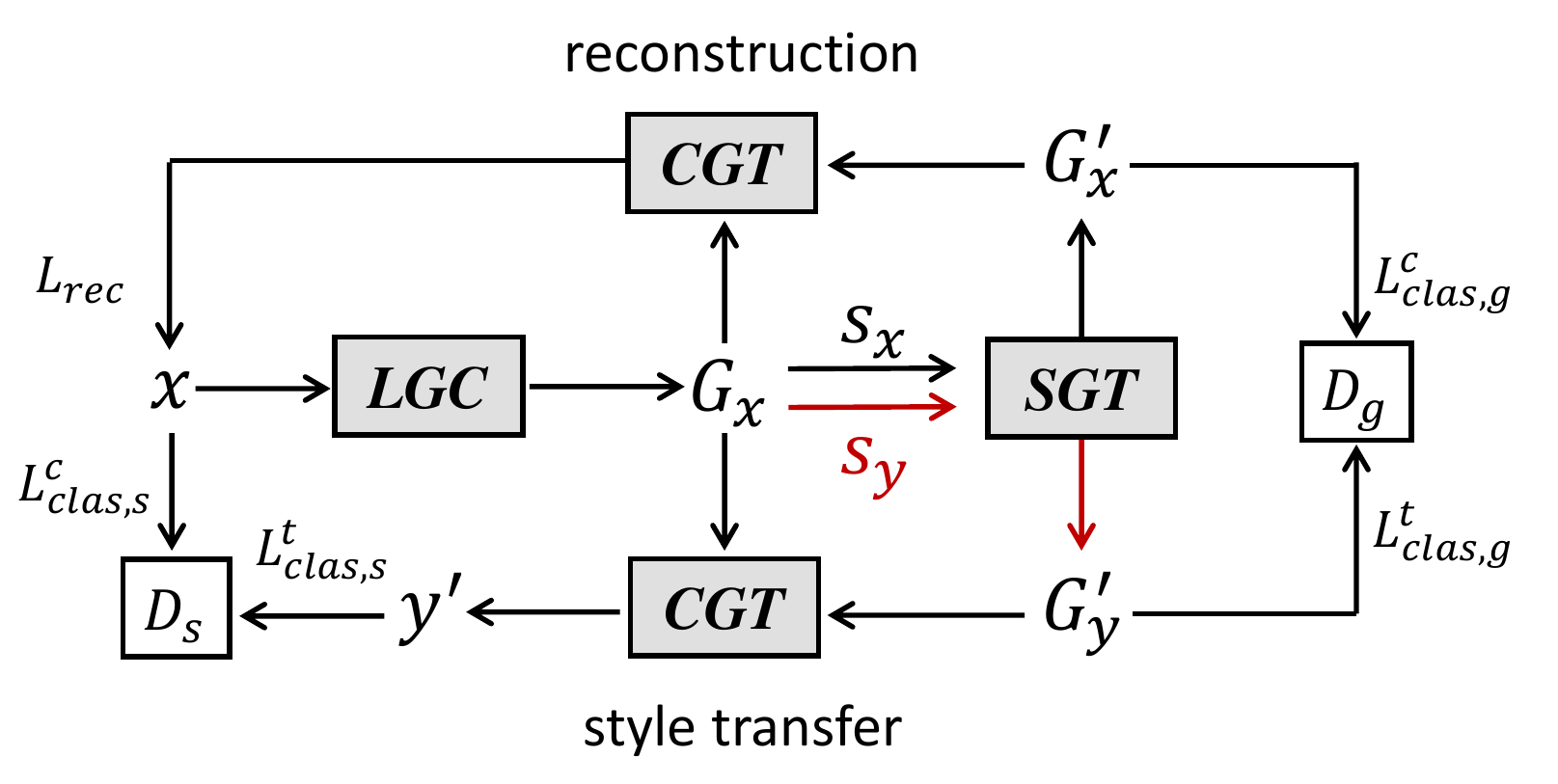}
  \caption{Implementation of the proposed Graph-Transformer based Auto Encoders (GTAE). The intrinsic linguistic graph $G_x$ is extracted by a linguistic graph constructor (LGC), followed by a self-graph transformer (SGT) with certain style attribute $\bm{s_x}/\bm{s_y}$ to obtain the latent graph embeddings $G'_x/G'_y$ . A cross-graph transformer (CGT) that takes the transferred graph $G'_x/G'_y$ and the original graph $G_x$ as input is then used for reconstruction with a rephraser. Two classifiers $\mathcal D_g$ and $\mathcal D_s$ are used at the graph and sentence levels respectively to back-propagate transfer signals.}
  \label{fig:overview}
  \vspace*{-2mm}
\end{figure}

{\color{black}{Equation~\ref{eq:framework} implies a graph-level encoder-decoder framework with an extra linguistic graph constructor (LGC), as depicted in Fig.~\ref{fig:overview}. In this work, we simplify the integral operations by modeling the three components as deterministic function.}} Specifically, we use a public available dependency parser for $p(G_x|\bm{x})$ to extract the linguistic graph of each sentence. A self-graph transformer (SGT) is proposed for the graph encoder $p(G'_y|G_x,\bm{s_y})$ to extract and transfer graph-level latent embeddings with the extracted linguistic graph and the desired style attribute. And we further design a cross-graph transformer (CGT) with a simple rephraser to model the graph decoder $p(\bm{y}|G'_y,G_x)$ which reconstructs the sentence from the transferred graph embeddings $G'_y$ while conditional on the original graph $G_x$. 

\subsection{Module Details}
\begin{figure*}[t]
  \centering
  \includegraphics[width=1.0\textwidth]{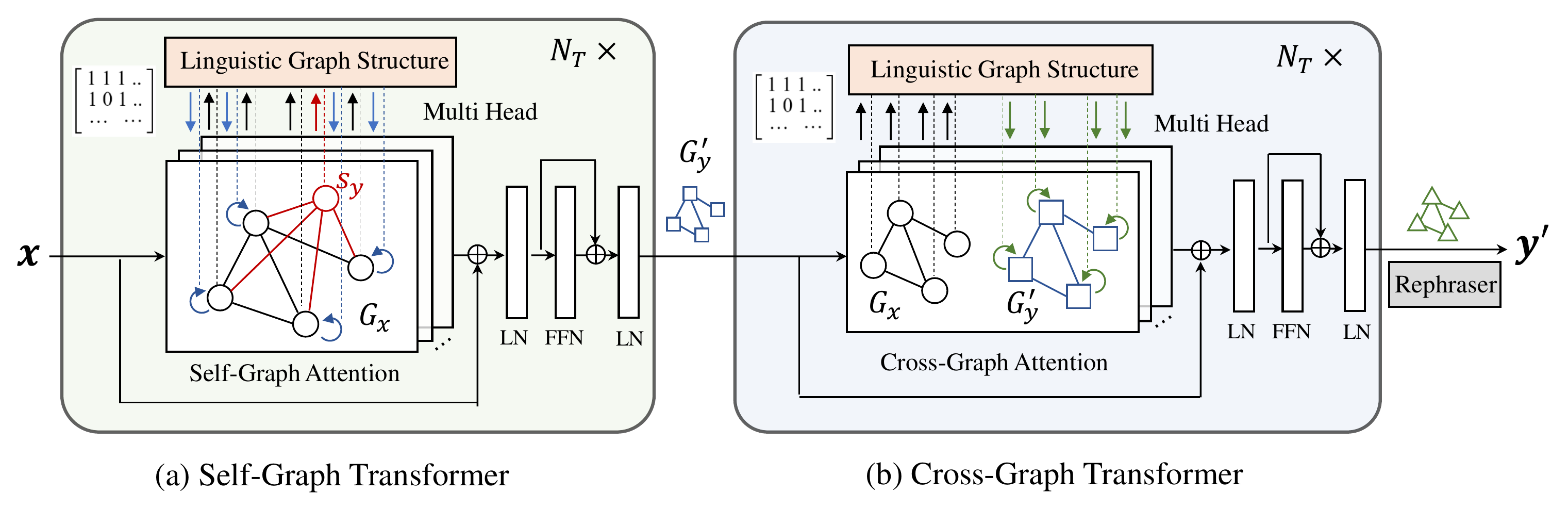}
  \caption{Model architectures of the proposed (a) self-graph transformer and (b) cross-graph transformer. Arrows between the attention module and the graph structure represent the information flow within/between the graphs constrained by the binarized linguistic adjacency matrix. LN and FFN denote layer normalization and feed-forward network respectively. And $N_T$ is the layer number.}
  \label{fig:graph_transformers}
  \vspace*{-2mm}
\end{figure*}

\paragraph{Linguistic Graph Constructor}
Due to the poor performance of current triplet-based information extraction systems for texts, we adopt a widely-used linguistic dependency parser~\cite{manning2014the} instead to construct the latent graph to represent the intrinsic content and logic of source sentences. For a sentence $\bm{x}=\{\bm{t}_i|i\in[1,k]\}$ with $k$ tokens, its corresponding linguistic graph $G_x$ can be represented as the set of the token-level nodes $V\in \mathbb R^{d_n\times k}$ and its adjacency matrix $E\in \mathbb R^{d_e\times k\times k}$ is built based on the grammar rules, where $d_n$ and $d_e$ are the feature size of node and edge embeddings respectively. {\color{black}{An illustration of the constructed linguistic graph is provided in Fig.~\ref{fig:illustration}.}} Since the goal of this work is to explore the research value of imposing such graph constraints rather than finetuning algorithmic details, we use a binarized version of the edges with $d_e=1$ currently for simplicity. Mapping the edges to a higher-level feature space and treating them differently in the following self/cross-graph transformers remain a future direction of this work.
\paragraph{Self-Graph Transformer}
\cite{li2018delete} pointed out that the style of a sentence is usually embodied in certain words or phrases. The procedure in their paper highly matches the graph structure proposed in our work, by deleting and retrieving certain nodes using a rule-based n-gram style classifier. In contrast to this hard-manipulation approach, our method let the graph nodes evolve progressively by attending to their neighbors and a global style node embedding, which offers more flexibility and better computation efficiency. In particular, inspired by recent success of attention-based graph neural networks, we propose a self-graph transformer (SGT) to encode $G_x = \{V_x, E_x\}$ given the specified style attribute, as depicted in Fig.~\ref{fig:graph_transformers}a. 

We basically follow the architecture setup of the transformer proposed in \cite{vaswani2017attention}. To incorporate the linguistic graph information, we restrict the attention mechanism to be effective only within the sets of connected graph nodes. Besides, a learnable global level style node $\bm{v'}\in\{\bm{s_x},\bm{s_y}\}$, which is connected to all nodes, is applied to provide style preservation or transfer signals during training and inference. {\color{black}{Specifically, we use a MLP layer to tranform the discrete style labels into a $d_n$-dimensional latent feature and set the edges between this style feature and the tokens to be 1.}} Suppose $V_x = \{\bm{v_x}^{(i)}|i\in[1,k]\}$ and $E_x = \{e_{ij}|i,j\in[1,k]\}$. At each layer, our graph attention with a residual function for each node reads:
\begin{eqnarray}
  &\bm{v_x}^{(i)} =  \bm{v_x}^{(i)} + \mathcal H\{\sigma[(W_q\bm{v_x}^{(i)})^T(W_kV_x^{(i)})/\sqrt{d_k}](W_vV_x^{(i)})\}, \\&V_x^{(i)} = \{\bm{v'},\bm{v_x}^{(j)}|e_{ij}>0\},
\end{eqnarray}
where $W_q,W_k\in \mathbb R^{d_k\times d_n}$ and $W_v\in \mathbb R^{d_v\times d_n}$ are the transformation matrices to map node embeddings into query, key and value embeddings respectively. $V_x^{(i)}\in \mathbb R^{d_n\times |V_x^{(i)}|}$ denotes the neighbor set of $\bm{v_x}^{(i)}$. $\sigma$ is a softmax function to normalize the attention weights and $\mathcal H$ is the multi-head attention function to learn and concatenate node representations in multiple subspaces. The proposed graph attention block is then followed by two layer-norm operations with a residual position-wise feed-forward network in between to extract higher level embeddings of each node.   

A graph-level style classifier $\mathcal D_g$ is adopted to transfer the node embeddings given the desired style at the graph level. We do not use the adversarial strategy due to its instability in the training phase. Instead, we train the classifier itself and its style transfer ability directly with the following two training objectives:
\begin{eqnarray}
  \mathcal L^c_{clas,g} = E_{G'_x\sim p(G'_x|G_x,\bm{s_x})}[-log\mathcal D_g(\bm{s_x}|G'_x)]\\ + E_{G'_y\sim p(G'_y|G_y,\bm{s_y})}[-log\mathcal D_g(\bm{s_y}|G'_y)],\\
  \mathcal L^t_{clas,g} = E_{G'_y\sim p(G'_y|G_x,\bm{s_y})}[-log\mathcal D_g(\bm{s_y}|G'_y)]\\ + E_{G'_x\sim p(G'_x|G_y,\bm{s_x})}[-log\mathcal D_g(\bm{s_x}|G'_x)]. 
\end{eqnarray}

\paragraph{Cross-Graph Transformer}
The graph decoder $p(\bm{y}|G'_y,G_x)$ derived in Equation~\ref{eq:framework} takes two graphs as input. An ideal $\bm{y'}$ generated from this distribution should borrow both the style information from $G'_y$ and the original content from $G_x$. To this end, we design a cross-graph transformer (CGT) to update node embeddings of $G'_y$ by attending to $G_x$ under the linguistic constraints, followed by a simple RNN rephraser to reconstruct the final transferred sentence, as illustrated in Fig.~\ref{fig:graph_transformers}b. 

The key difference that distinguishes our proposed CGT from other counterparts is that a cross-graph attention mechanism is applied to allow information flow from the source graph for better semantic and logic reconstruction. Assuming $G'_y$ and $G_x$ share the same linguistic graph structure, we have $G'_y = \{V'_y,E_x\}$ with $V'_y = \{\bm{{v'_y}}^{(i)}|i\in [1,k]\}$. Then the formula of cross-graph attention reads:
\begin{equation}
\begin{split}
  \bm{{v'_y}}^{(i)} =& \\ \bm{{v'_y}}^{(i)}  +& \mathcal H\{\sigma[(W'_q\bm{{v'_y}}^{(i)})^T(W'_kV_x^{(i)})/\sqrt{d_k}](W'_vV_x^{(i)})\}, V_x^{(i)} \\&= \{\bm{v_x}^{(j)}|e_{ij}>0\}.
\end{split}
\end{equation}

The reconstruction loss for the proposed graph-level encoder-decoder is:
\begin{eqnarray}
  \mathcal L_{rec} = E_{G'_x\sim p(G'_x|G_x,\bm{s_x})}[-logp(\bm{x}|G'_x,G_x)] \\ + E_{G'_y\sim p(G'_y|G_y,\bm{s_y})}[-logp(\bm{y}|G'_y,G_y)].
\end{eqnarray}

To ensure the style transfer intensity after reconstruction, we further adopt a sentence-level style classifier $\mathcal D_s$, trained in the same way as $\mathcal D_g$ with the following objectives:
\begin{eqnarray}
  \mathcal L^c_{clas,s} & = & E_{\bm{x}\sim \bm{X}}[-log\mathcal D_s(\bm{s_x}|\bm{x})] \\ &+& E_{\bm{y}\sim \bm{Y}}[-log\mathcal D_s(\bm{s_y}|\bm{y})],\\
  \mathcal L^t_{clas,s} & = & E_{\bm{y'}\sim p(\bm{y}|G'_y,G_x)}[-log\mathcal D_s(\bm{s_y}|\bm{y'})] \\& +& E_{\bm{x'}\sim p(\bm{x}|G'_x,G_y)}[-log\mathcal D_s(\bm{s_x}|\bm{x'})].
\end{eqnarray}

\paragraph{Training Scheme}
Our proposed GTAE is trained in an end-to-end manner. Gumbel-softmax continuous approximation is adopted in the RNN rephraser to overcome the discreteness problem of text outputs. We first pretrain the model for a few warm-up epochs by iteratively backpropagating the reconstruction loss $\mathcal L_{rec}$ and the joint classification loss $\mathcal L^c_{clas,g}+\mathcal L^c_{clas,s}$. {\color{black}{Then in the training phase, we instead iteratively backpropagate the transfer loss $\mathcal L_{rec} + \lambda_g \mathcal L^t_{clas,g} + \lambda_s \mathcal L^t_{clas,s}$ for training the SGT, CGT and rephraser modules and the classification loss $\mathcal L^c_{clas,g}+\mathcal L^c_{clas,s}$ for further training the classifiers. $\lambda_g$ and $\lambda_s$ are the hyperparameters that control the overall transfer strength. It is worth noting that though we consider binary-style transfer in this paper, our method can be directly extended to multi-style transfer, where we can add corresponding classification and reconstruction losses to $\mathcal L^c_{clas,g}$, $\mathcal L^c_{clas,s}$, and $\mathcal L_{rec}$. For the transfer losses, we can either consider adding combinatorial transfer losses into $\mathcal L^t_{clas,g}$ and $\mathcal L^t_{clas,s}$, or iteratively select two styles and conduct training in the same manner as the binary case.}}

\begin{table}[t]
	\centering
	\caption{Statistics of the three datasets}
	\resizebox{0.45\textwidth}{13mm}{
	\begin{tabular}{c|c|c|c}
		\hline
		Dataset & Style & Size & Vocab \\ 
		\hline \hline
		\multirow{2}*{\centering Yelp Sentiment} & positive & 382,917 & \multirow{2}*{\centering 9,357}\\ 
		~  & negative & 256,026 & ~ \\
		\hline
		\multirow{2}*{\centering Political Slant} & democratic & 170,423 & \multirow{2}*{\centering 11,441}\\ 
		~  & republican & 176,011 & ~ \\
		\hline
		\multirow{2}*{\centering Paper-News Title} & papers & 95,905 & \multirow{2}*{\centering 19,072}\\ 
		~  & news & 104,633 & ~ \\
		\hline
	\end{tabular}}
	
	\label{tb:data_stat}
\end{table}

\section{Experiments}

We verify the effectiveness of our proposed method on three non-parallel text style transfer datasets: yelp restaurant reviews, political slant comments and paper-news titles. Following a recent work on text style transfer evaluation~\cite{mir2019evaluating}, we compare our approach with three representative baselines~\cite{shen2017style, zhao2018adversarially, li2018delete}, i.e. the cross-aligned auto-encoders (CAAE), the adversarially regularized autoencoders (ARAE) and the delete-retrive-generate model (DAR) across all the three datasets. {\color{black}{For the most representative yelp dataset w.r.t. sentiment transfer, we further conducted a more thorough comparison with a list of the state-of-the-arts methods.}} Our proposed GTAE achieves the best content preservation ability as expected, while maintaining highly comparable performance on naturalness and style transfer intensity with the state-of-the-arts. The model is implemented using the Texar~\cite{hu2018texar} tookit for text generation based on the Tensorflow backend~\cite{abadi2016tensorflow}.

\subsection{Datasets}

For sentiment transfer task, we use the Yelp restaurant reviews dataset released in \cite{shen2017style}. The political slant dataset is retrieved from top-level comments on Facebook posts from the members of United States Senate and House~\cite{prabhumoye2018style}. We also report results on the dataset of paper-news title types released in \cite{fu2018style}. Following \cite{shen2017style}, we filter out sentences with length larger than 15 and replace words that occurs less than 5 times with <unk>. The datasets are randomly splited into train/dev/test splits in the same manner as original papers. Summary statistics of the datasets are listed in Table~\ref{tb:data_stat}.

\subsection{Evaluation Metrics}
A successful text style transfer model normally requires the capacities in three aspects: content preservation, style transfer intensity and naturalness. Content preservation aims at minimizing the semantic information loss of style-irrelevant components of original sentences, while style transfer intensity is used to measure the correctness of transferred sentences in terms of the given styles. To ensure that the generated sentences should be human-readable and avoids looking like machine-generated sentences, the evaluation of sentence naturalness is also required. Due to the fact that human evaluation is quite labor-intensive and time-consuming, automatic evaluation of the above aspects is desired for comparison of the methods in this task. 

However, to the best of our knowledge, there is no consensus on automatic evaluation metrics for text style transfer. The main difficulty lies in the lack of parallel corpus. Besides, even if we have labelled data, direct comparison between generated results and human-transferred sentences is still an ill-posed problem since there usually exist multiple ways to express a sentence with the specific style. Recently, \cite{mir2019evaluating} systematically studied this problem by evaluating possible metrics for text style transfer based on the correlation coefficients with human ratings. They proposed to use EMD~\cite{rubner1998a, pele2009fast}, WMD~\cite{kusner2015from} and adversarial naturalness for the three aspects mentioned above. In this work, we basically follow the practice in \cite{mir2019evaluating} and report these metrics in the results.

\paragraph{Style Transfer Intensity}
To determine the degree to which a sentence is succesfully transferred, previous works relied on a pre-trained classifier to calculate the empirical transfer accuracy. However, the binarized classifier used in this strategy could miss the subtle difference between transferred sentences when they exhibit varied transfer intensities but labelled with the same value without discrimination. This problem was overcome in EMD by measuring the style distribution difference instead, which also leads to WMD for content preservation.
\paragraph{Content Preservation}
BLEU~\cite{papineni2002bleu} is probably the most popular metric for content preservation evaluation by measuring the n-gram similarity w.r.t. original sentences. \cite{li2018delete} proposed to calculate BLEU w.r.t. human-transferred sentences instead and released small test datasets with human labels for that purpose. Nevertheless, the fact that possible transferred sentences are usually not unique and the high cost of human labors limit the usage of their method. WMD provides a more dedicated distribution-based metric by calculating the minimum distance between word embeddings. To remove style-relevant parts of the sentences to compare, we build a style lexicon by training a regression model and mask style-related words in WMD calculation as suggested by \cite{mir2019evaluating}. In addition, we report a recently published metric BERTSCORE~\cite{zhang2019bertscore} which claims higher-level semantic alignments as an alternative measurement for comparison.
\paragraph{Naturalness}
Perplexity calculated from pre-trained language models provides a solution for naturalness measurement, but still it is not necessarily a gold standard. The adversarial naturalness proposed in \cite{mir2019evaluating} trains a classifier to distinguish human-generated texts from machine-generated texts first. Then naturalness reduces to the ability to fool this classifier. \cite{mir2019evaluating} showed that adversarial naturalness serves as a better measurement in terms of their correlation coefficient w.r.t. human evaluation. In our experiments, we use the three pre-trained classifiers released in \cite{mir2019evaluating} and report the corresponding adversarial naturalness scores respectively.
\subsection{Experimental Settings}
We use the StanfordCoreNLP toolbox~\cite{manning2014the} to extract the linguistic structure of each sentence. For the SGT and CGT modules, the numbers of heads and layers are 8 and 2 respectively. We choose 512 as the size of all hidden embeddings. The rephraser is implemented as a single-layer GRU encoder-decoder for final sentence generation. And the TextCNN model~\cite{kim2014convolutional} is used for both classifiers. We pretrain the model for 10 epochs, followed by two or three extra training epochs, depending on validation errors. We empirically choose $\lambda_g$ and $\lambda_s$ as $0.05$ and $0.02$ respectively. We set the temperature anneal rate for Gumbel-softmax approximation as 0.5. The model is trained using Adam optimizer~\cite{kingma2015adam} with an initial learning rate 5e-4. Batch size is empirically set to be 128 for all the three datasets. We report the results on randomly selected 1000 test samples due to the heavy computational cost of the retrival process in DAR. {\color{black}{All the training and evaluation source codes as well as the results are released in this page~\footnote{https://github.com/SenZHANG-GitHub/GTAE}.}}

\begin{table*}[t]
	\centering
	\caption{Sentiment transfer results on the test set.} 
	
	\begin{tabular}{c | c c c c c | c c c | c c}
		\hline
		\multirow{2}*{Model} & \multicolumn{5}{c |}{Content Preservation} & \multicolumn{3}{c |}{Naturalness} & \multicolumn{2}{c}{Transfer Intensity} \\
		\cline{2-11}
		~ & mWMD & BLEU & B-P & B-R & B-F1 & N-A & N-C & N-D & ACCU & EMD \\ 
		\hline \hline
		\textbf{GATE (ours)}         & \textbf{0.1027} & \textbf{64.83} & 0.6991 & \textbf{0.7303} & 0.7149 & 0.6178 & 0.9272 & 0.6644 & 0.8870    & 0.8505    \\ 
		StyleTrans-multi~\cite{dai2019style} & 0.1536          & 63.08          & \textbf{0.7145} & 0.7203          & \textbf{0.7175}          & 0.6133          & 0.9102          & 0.6909          & 0.8730             & 0.8316  \\
		DualRL~\cite{luo2019dual}                 & 0.1692          & 59.01          & 0.7125          & 0.6988          & 0.7057          & 0.5517          & 0.8996          & 0.6768          & 0.9050             & 0.8675             \\ 
		StyleTrans-cond~\cite{dai2019style}  & 0.2223          & 53.28          & 0.6205          & 0.6475          & 0.6341          & 0.6312          & 0.9109          & 0.6654          & 0.9290             & 0.8815             \\ 
		UnsuperMT~\cite{zhang2018style}       & 0.2450          & 46.25          & 0.6060          & 0.6206          & 0.6134          & 0.5755          & 0.9040          & 0.6625          & \textbf{0.9770}    & \textbf{0.9372}    \\ 
		UnpairedRL~\cite{xu2018unpaired}         & 0.3122          & 46.09          & 0.4504          & 0.4709          & 0.4612          & 0.7136          & 0.9035          & 0.6493          & 0.5340             & 0.4989             \\ 
		DAR\_Template~\cite{li2018delete}       & 0.4156          & 57.10          & 0.4970          & 0.5406          & 0.5185          & 0.6370          & 0.8984          & 0.6299          & 0.8410             & 0.7948             \\ 
		DAR\_DeleteOnly~\cite{li2018delete}         & 0.4538          & 34.53          & 0.4158          & 0.4823          & 0.4490          & 0.6345          & 0.9072          & 0.5511          & 0.8750             & 0.8297             \\ 
		DAR\_DeleteRetrieve~\cite{li2018delete}     & 0.4605          & 36.72          & 0.4268          & 0.4799          & 0.4534          & 0.6564          & 0.9359          & 0.5620          & 0.9010             & 0.8550             \\ 
		CAAE~\cite{shen2017style}             & 0.5130          & 20.74          & 0.3585          & 0.3825          & 0.3710          & 0.4139          & 0.7006          & 0.5999          & 0.7490             & 0.7029             \\ 
		IMaT~\cite{jin2019imat}                   & 0.5571          & 16.92          & 0.4750          & 0.4292          & 0.4501          & 0.4878          & 0.8407          & 0.6691          & 0.8710             & 0.8198             \\ 
		Multi\_Decoder~\cite{fu2018style}     & 0.5799          & 24.91          & 0.3117          & 0.3315          & 0.3223          & 0.4829          & 0.8394          & 0.6365          & 0.6810             & 0.6340             \\ 
		FineGrained-0.7~\cite{luo2019towards}       & 0.6239          & 11.36          & 0.4023          & 0.3404          & 0.3717          & 0.3665          & 0.7125          & 0.5332          & 0.3960             & 0.3621             \\ 
		FineGrained-0.9~\cite{luo2019towards}        & 0.6251          & 11.07          & 0.4030          & 0.3389          & 0.3713          & 0.3668          & 0.7148          & 0.5231          & 0.4180             & 0.3926             \\ 
		FineGrained-0.5~\cite{luo2019towards}        & 0.6252          & 11.72          & 0.3994          & 0.3436          & 0.3718          & 0.3606          & 0.7254          & 0.5395          & 0.3280             & 0.2895             \\
		BackTranslation~\cite{prabhumoye2018style}    & 0.7566          & 2.81           & 0.2405          & 0.2024          & 0.2220          & 0.3686          & 0.5392          & 0.4754          & 0.9500             & 0.9117             \\ 
		Style\_Emb~\cite{fu2018style}       & 0.8796          & 3.24           & 0.0166          & 0.0673          & 0.0429          & 0.5788          & 0.9075          & 0.6450          & 0.4490             & 0.4119             \\ 
		DAR\_RetrieveOnly~\cite{li2018delete}       & 0.8990          & 2.62           & 0.1368          & 0.1818          & 0.1598          & \textbf{0.8067} & \textbf{0.9717} & \textbf{0.7211} & 0.9610             & 0.9010             \\ 
		ARAE~\cite{zhao2018adversarially}                   & 0.9047          & 5.95           & 0.1680          & 0.1478          & 0.1584          & 0.4476          & 0.8120          & 0.6969          & 0.8278             & 0.7880             \\ \hline
	\end{tabular}
	
	\label{tb:yelp_result}
\end{table*}

  
\subsection{Results}
This section shows the experiment results in the perspective of evaluation metrics. Specifically, in Table~\ref{tb:yelp_result}, WMD is the masked Word Mover's distance with style-related tokens removed, which serves as the major metric for content preservation evaluation. B-P, B-R and B-F1 denote the precision, recall and F1 score of the BERTSCORE, which are rescaled as suggested in [] for better illustration. {\color{black}{We also report the BLEU scores between the original and transfer sentences.}} A smaller WMD means the transfer sentence is closer to the original one, while higher BLEU and BERTSCORE metrics indicate better content preservation ability. N-A, N-C and N-D are the adversarial naturalness scores calculating from the three released classifiers trained on ARAE, CAAE and DAR respectively, where a higher score means a higher probability to be "natural"~\cite{mir2019evaluating}. ACCU and EMD denote empirical transfer accuracy and Earth Mover's Distance based on a pretrained CNN classifier. Higher ACCU and EMD scores indicate a stronger transfer intensity. {\color{black}{For the yelp dataset, we evaluated the results released in \cite{luo2019dual} for DualRl~\cite{luo2019dual}, UnsuperMT~\cite{zhang2018style}, UnpairedRL~\cite{xu2018unpaired}, and DAR-related methods~\cite{li2018delete}, the results released in \cite{jin2019imat} for IMaT~\cite{jin2019imat}, CAAE~\cite{shen2017style}, Multi\_Decoder~\cite{fu2018style}, and Style\_Emb~\cite{fu2018style}, and the results of StyleTrans~\cite{dai2019style} and FineGrained~\cite{luo2019towards} methods that are released on their own websites. Note that for the FineGrained results, the authors only released a subset that containing 500 rather than 1000 sentences. We also trained ARAE~\cite{zhao2018adversarially} under its default setting on the yelp dataset. For the political and title datasets, we trained CAAE, ARAE and DAR-related methods using their released source codes under the default settings as well.}}

\paragraph{Yelp Sentiment Dataset}

\begin{figure*}
  \centering
  \includegraphics[width=\textwidth]{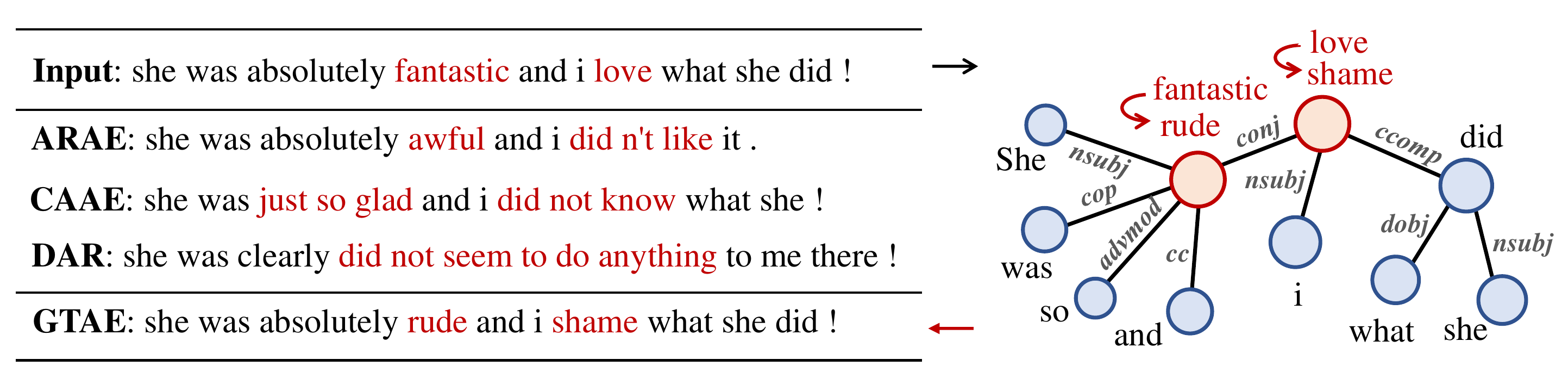}
  \caption{An example of sentiment transfer results. The style is supposed to be transferred from positive to negative. By leveraging the established linguistic graph, our model identifies and transfers the style-related graph nodes while preserving the overall graph structure.}
  \label{fig:example}
\end{figure*}

{\color{black}{Table~\ref{tb:yelp_result} shows the evaluation results on the Yelp sentiment dataset, where the methods are ranked by the masked WMD metric (The lower the better). As expected, our model achieves the best content preservation performance in terms of the masked WMD, BLEU, and BERT-Recall scores, while maintaining a reasonable transfer intensity and naturalness scores. It might be not surprising that retrieval-based DAR achieves the best naturalness scores and also a high transfer intensity since real human-written sentences are retrieved as results. However, the retrieval based strategies sacrifice a lot w.r.t. content preservation ability and take around five seconds for the extra retrieval process during inference, which is practically unacceptable for real-world applications. UnsuperMT achieves the best transfer intensity scores, but also at the price of content preservation ability and language naturalness. In terms of content preservation, style transformer with multi-class discriminator is the one that is closest to our method, while our model outperforms this method w.r.t. naturalness and transfer intensity.}} It can also be seen that the difference of EMD could be smaller than that of transfer accuracy, supporting that EMD serves as a more dedicated metric by leveraging the style distribution information. Fig.~\ref{fig:example} gives an example of sentiment transfer results and illustrates how our model works by leveraging the linguistic graph.

\paragraph{Political Slant Dataset}
The evaluation results on the political slant dataset are similar to sentiment transfer results as shown in Table~\ref{tb:political_result}. Our proposed GTAE still performs significantly better than others in preserving the style-irrelevant contents while maintaining competitive results on naturalness and transfer intensity. Besides, the unsatisfatory content preservation capbility of ARAE might result from the unstable adversarial training process. In comparison, our method avoids the adversarial strategy by directly back-propagating style transfer signals at both graph and sentence levels. 
\begin{table*}
	\centering
	\caption{Political slant transfer results on the test set.} 
	\resizebox{\textwidth}{15mm}{
	\begin{tabular}{c | c c c c c | c c c | c c}
		\hline 
		\multirow{2}*{Model} & \multicolumn{5}{c |}{Content Preservation} & \multicolumn{3}{c |}{Naturalness} & \multicolumn{2}{c}{Transfer Intensity} \\
		\cline{2-11}
		~ & WMD & BLEU & B-P & B-R & B-F1 & N-A & N-C & N-D & ACCU & EMD \\ 
		\hline \hline
		\textbf{GTAE (ours)}  & \textbf{0.1506} & \textbf{65.61} & \textbf{0.6577} & \textbf{0.6706} & \textbf{0.664} & \textbf{0.3310} & 0.7852          & 0.7318          & 0.900              & 0.8990             \\ 
		CAAE~\cite{shen2017style}                   & 0.4968          & 15.68          & 0.3217          & 0.3240          & 0.3230         & 0.2715          & 0.7052          & \textbf{0.7370} & 0.828              & 0.8259             \\ 
		DAR\_DeleteOnly~\cite{li2018delete}        & 0.5000          & 30.76          & 0.3295          & 0.3932          & 0.3605         & 0.3155          & \textbf{0.8534} & 0.6490          & 0.958              & 0.9565             \\ 
		DAR\_DeleteRetrieve~\cite{li2018delete}                   & 0.5109          & 35.48          & 0.3352          & 0.3966          & 0.3649         & 0.3190          & 0.8472          & 0.7081          & 0.977              & 0.9747             \\ 
		DAR\_RetrieveOnly~\cite{li2018delete}      & 0.7771          & 10.14          & 0.1590          & 0.1840          & 0.1709         & 0.3219          & 0.7854          & 0.7271          & \textbf{0.998}     & \textbf{0.9960}    \\ 
		ARAE~\cite{zhao2018adversarially}                   & 1.0347          & 2.95           & 0.0203          & 0.0117          & 0.0158         & 0.3092          & 0.7763          & 0.7333          & 0.944              & 0.9412 \\ \hline
	\end{tabular}}
	
	\label{tb:political_result}
\end{table*}

\begin{table*}[ht!]
	\centering
	\caption{Paper-news title transfer results on the test set.} 
	\resizebox{\textwidth}{15mm}{
	\begin{tabular}{c | c c c c c | c c c | c c}
		\hline
		\multirow{2}*{Model} & \multicolumn{5}{c |}{Content Preservation} & \multicolumn{3}{c |}{Naturalness} & \multicolumn{2}{c}{Transfer Intensity} \\
		\cline{2-11}
		~ & WMD & BLEU & B-P & B-R & B-F1 & N-A & N-C & N-D & ACCU & EMD \\ 
		\hline \hline
		\textbf{GTA (ours)} & \textbf{0.5161} & \textbf{19.67} & \textbf{0.1923} & \textbf{0.2134} & \textbf{0.2034} & 0.0949          & 0.4181          & 0.4567          & 0.956              & 0.9492             \\ 
		DAR\_DeleteOnly~\cite{li2018delete}      & 0.8413          & 4.75           & 0.0939          & 0.0412          & 0.0677          & \textbf{0.3912} & \textbf{0.8374} & 0.4495          & 0.881              & 0.8687             \\ 
		DAR\_DeleteRetrieve~\cite{li2018delete}  & 0.8567          & 5.04           & 0.0249          & 0.0212          & 0.0234          & 0.2462          & 0.7387          & 0.4625          & 0.933              & 0.9234             \\ 
		CAAE~\cite{shen2017style}                 & 0.9226          & 0.82           & 0.0067          & -0.0099         & -0.0008         & 0.2167          & 0.5627          & 0.4422          & 0.972              & 0.9612             \\ 
		DAR\_RetrieveOnly~\cite{li2018delete}    & 0.9842          & 0.37           & -0.0383         & -0.0362         & -0.0365         & 0.1490          & 0.5701          & 0.4261          & \textbf{0.995}     & \textbf{0.9856}    \\ 
		ARAE~\cite{zhao2018adversarially}                 & 1.0253          & 0.00           & -0.0447         & -0.0539         & -0.0486         & 0.2318          & 0.6061          & \textbf{0.4765} & 0.989              & 0.9782             \\ \hline
	\end{tabular}}
	\label{tb:title_result}
\end{table*}

\paragraph{Paper-News Title Dataset} 

This task is more challenging due to the larger semantic domain mis-alignment between paper and news titles, which explains the worse content preservation capbilities of all models as shown in Table~\ref{tb:title_result}. Moreover, since the released adversarial naturalness classifiers are trained on Yelp reviews data, the domain mis-alignment between online reviews and formal titles also leads to a reduction of naturalness scores. Our model still reaches the best content preservation performance over the three baselines, with comparable naturalness and transfer intensity. 

{\color{black}{\subsection{Ablation Analysis}
We further examine the effect of each component in proposed model on the yelp sentiment dataset.

\paragraph{Effect of the Linguistic Graph Constraint}
We first examine the effect of the linguistic graph constraint by setting the adjacency matrix used in SGT and CGT to the identity matrix, leading to the models SGT-I and CGT-I respectively. The model that sets both adjacency matrices in SGT and CGT to identity matrix is also evaluated (SGT-CGT-I). We report the results in Tabel~\ref{tb:abl_I}. Without the linguistic graph constraint, both content preservation and transfer intensity abilities degrade. The linguistic constraints in SGT and CGT are more crucial for content preservation and style transfer respectively.}}

\begin{table*}[ht!]
	\centering
	\caption{{\color{black}Ablation results of the linguistic graph on the yelp dataset.} }
	\resizebox{0.95\textwidth}{12mm}{
	\begin{tabular}{c | c c c c c | c c c | c c}
		\hline
		\multirow{2}*{Model} & \multicolumn{5}{c |}{Content Preservation} & \multicolumn{3}{c |}{Naturalness} & \multicolumn{2}{c}{Transfer Intensity} \\
		\cline{2-11}
		~ & WMD & BLEU & B-P & B-R & B-F1 & N-A & N-C & N-D & ACCU & EMD \\ 
		\hline \hline
		GTAE-full & 0.1027       & 64.83         & 0.6991       & 0.7303       & 0.7149        & 0.6178       & 0.9272       & 0.6644       & 0.887               & 0.8505            \\ 
		SGT-I & 0.1394       & 61.10         & 0.6591       & 0.6735       & 0.6667        & 0.6692       & 0.9313       & 0.7141       & 0.881               & 0.8437            \\
		CGT-I & 0.1211       & 62.71         & 0.6772       & 0.6982       & 0.6878        & 0.6347       & 0.9297       & 0.6164       & 0.871               & 0.8368            \\ 
		SGT-CGT-I & 0.1369       & 64.77         & 0.6809       & 0.7055       & 0.6935        & 0.7509       & 0.9681       & 0.7193       & 0.843               & 0.8038            \\ \hline
	\end{tabular}}
	\label{tb:abl_I}
\end{table*}

{\color{black}{\paragraph{Effect of the Graph- and Sentence-level Losses}
We next study the effect of the graph- and sentence-level losses. We remove $\mathcal L^c_{clas,s}$ and $\mathcal L^c_{clas,g}$ when training the classifiers, leading to two models c-clas-g-only and c-clas-s-only respectively. Besides, we also We remove $\mathcal L^t_{clas,s}$ and $\mathcal L^t_{clas,g}$ when training the style transfer ability, leading to the models t-clas-g-only and t-clas-s-only, respectively. Results are given in Table~\ref{tb:abl_loss}, which show that a properly trained sentence-level classifier is crucial for the style transfer ability. Moreover, the transfer signals from both graph- and sentence-level are both important for achieving a good style transfer performance.}}

\begin{table*}[ht!]
	\centering
	\caption{{\color{black}Ablation results of the graph- and sentence- level losses on the yelp dataset.} }
	\resizebox{0.95\textwidth}{14mm}{
	\begin{tabular}{c | c c c c c | c c c | c c}
		\hline
		\multirow{2}*{Model} & \multicolumn{5}{c |}{Content Preservation} & \multicolumn{3}{c |}{Naturalness} & \multicolumn{2}{c}{Transfer Intensity} \\
		\cline{2-11}
		~ & WMD & BLEU & B-P & B-R & B-F1 & N-A & N-C & N-D & ACCU & EMD \\ 
		\hline \hline
		GTAE-full & 0.1027       & 64.83         & 0.6991       & 0.7303       & 0.7149        & 0.6178       & 0.9272       & 0.6644       & 0.887               & 0.8505            \\ 
		c-clas-g-only & 0.0017       & 99.64         & 0.9970       & 0.9969       & 0.9970        & 0.6738       & 0.9357       & 0.6841       & 0.027               & -0.0003           \\ 
		c-clas-s-only & 0.1424       & 61.64         & 0.6704       & 0.6927       & 0.6818        & 0.6586       & 0.9217       & 0.7069       & 0.854               & 0.8159            \\ 
		t-clas-g-only & 0.0016       & 99.63         & 0.9963       & 0.9963       & 0.9963        & 0.6742       & 0.9358       & 0.6845       & 0.028               & 0.0008    \\
		t-clas-s-only & 0.0106       & 92.83         & 0.9480       & 0.9503       & 0.9492        & 0.6317       & 0.9278       & 0.6933       & 0.203               & 0.1725        \\ \hline 
	\end{tabular}}
	\label{tb:abl_loss}
\end{table*}

{\color{black}{\paragraph{Effect of the Warm-up Phase}
The warm-up phase prepares pretrained graph- and sentence- level classifiers and a basic reconstruction ability for the following training phase. We study the effect of the pretrain epochs in this section and summarize the results in Table~\ref{tb:abl_pretrain}. While content preservation ability is relatively robust to the pretrain epochs, a certain level of pretraining is essential to achieve a satisfactory transfer intensity performance.}}

\begin{table*}[ht!]
	\centering
	\caption{{\color{black}Ablation results of the pretrain epochs on the yelp dataset.} }
	\resizebox{0.95\textwidth}{15mm}{
	\begin{tabular}{c | c c c c c | c c c | c c}
		\hline
		\multirow{2}*{Model} & \multicolumn{5}{c |}{Content Preservation} & \multicolumn{3}{c |}{Naturalness} & \multicolumn{2}{c}{Transfer Intensity} \\
		\cline{2-11}
		~ & WMD & BLEU & B-P & B-R & B-F1 & N-A & N-C & N-D & ACCU & EMD \\ 
		\hline \hline
		pretrain-0 & 0.0999       & 67.43         & 0.7252       & 0.7305       & 0.7282        & 0.6288       & 0.9261       & 0.7282       & 0.798               & 0.7618            \\ 
		pretrain-2 & 0.0943       & 66.32         & 0.7178       & 0.7275       & 0.7229        & 0.6312       & 0.9341       & 0.6927       & 0.849               & 0.8139            \\ 
		pretrain-4 & 0.1598       & 60.20         & 0.6439       & 0.6749       & 0.6596        & 0.6840       & 0.9259       & 0.6424       & 0.878               & 0.8377            \\ 
		pretrain-6 & 0.1093       & 64.87         & 0.6957       & 0.7150       & 0.7057        & 0.6397       & 0.9580       & 0.6987       & 0.855               & 0.8160            \\ 
		pretrain-8 & 0.1152       & 67.00         & 0.7017       & 0.7225       & 0.7124        & 0.6645       & 0.9529       & 0.6821       & 0.836               & 0.7980            \\ 
		pretrain-10  & 0.1027       & 64.83         & 0.6991       & 0.7303       & 0.7149        & 0.6178       & 0.9272       & 0.6644       & 0.887               & 0.8505            \\ \hline 
	\end{tabular}}
	\label{tb:abl_pretrain}
\end{table*}

\section{Conclusion}
In this work, we propose to impose intrinsic linguistic constraints into the model space of natural language to achieve a better content and logic preservation ability for non-parallel text style transfer. We formulate this procedure as a graph-level encoder-decoder process and propose a Graph-Transformer based Auto-Encoder (GTAE) with two novel modules, namely Self-Graph Transformer (SGT) and Cross-Graph Transformer (CGT) for better manipulation of the latent graph embeddings. Experiments on three non-parallel text style transfer datasets demonstrate that our model achieves the best content perservation performation, while maintaining state-of-the-art transfer intensity and naturalness. Besides, our proposed GTAE is highly compatible with other methods. Potential future works include incorporating language models as discriminators to improve the naturalness and developing methods that allow varied edge types to better leverage the linguistic structures. Moreover, extending this idea into more generic text generation tasks is also a promising and exciting research direction. 
\ifCLASSOPTIONcaptionsoff
  \newpage
\fi

\bibliographystyle{IEEEtran}
\bibliography{egbib}




\end{document}